\documentclass[sigconf]{acmart}
\authorsaddresses{}
\AtBeginDocument{%
  }

\usepackage{xcolor}
\usepackage{acronym}
\usepackage{enumitem}
\usepackage{graphicx}
\usepackage{subcaption}
\usepackage{colortbl}
\usepackage{siunitx}
\definecolor{best}{RGB}{255, 200, 200}   % light red/pink  — best value
\definecolor{second}{RGB}{242, 242, 200} % light yellow    — second best
 
\acmDOI{3799825.3818700}
\copyrightyear{2026}
\acmYear{2026}
\setcopyright{cc}
\setcctype{by}
\acmConference[SIGGRAPH Posters '26]{Special Interest Group on Computer Graphics and Interactive Techniques Conference Posters}{July 19--23, 2026}{Los Angeles, CA, USA}
\acmBooktitle{Special Interest Group on Computer Graphics and Interactive Techniques Conference Posters (SIGGRAPH Posters '26), July 19--23, 2026, Los Angeles, CA, USA}
\acmDOI{10.1145/3799825.3818700}
\acmISBN{979-8-4007-2548-7/2026/07}
\newacro{CGVQ}{Cluster-Guided Vector Quantization}
\newacro{GS}{Gaussian Splatting}
\newacro{GI}{GaussianImage}
\newacro{bpp}{Bit-Per-Pixel}
\newcommand{\argmin}{\mathop{\mathrm{argmin}}}
\newcommand{\refFig}[1]{Fig.~\ref{fig:#1}}

\begin{document}

\title{Clustered Codebook Quantization for 2D Gaussian-based Image Compression}

\author{Runze Cheng}
\authornote{Primary contributor to this work}
\orcid{0009-0000-2629-9635}

\affiliation{%
  \institution{University College London}
  \city{London}
  \country{UK}
}

\author{Yicheng Zhan}

\affiliation{%
  \institution{University College London}
  \city{London}
  \country{UK}
}

\author{Josef Spjut}

\affiliation{%
  \institution{NVIDIA}
  \city{Durham}
  \country{USA}
}

\author{Kaan Akşit}

\affiliation{%
  \institution{University College London}
  \city{London}
  \country{UK}
}

\renewcommand{\shortauthors}{Cheng et al.}

\begin{abstract}
Gaussian-based image representations effectively model image content using compact parametric primitives while preserving high visual fidelity \cite{gsImage}, yet storing a large number of floating-point parameters per primitive degrades rate-distortion efficiency at higher fidelity targets.
To improve the rate-distortion performance in Gaussian representation, we present our \textcolor{blue}{\textbf{\ac{CGVQ}}}, a Gaussian primitive based image compression method.
Our key idea is to partition Gaussian parameters further into homogeneous groups prior to quantization, enabling higher compression efficiency and accurate parameter reconstruction.  
In practice, our extensive experiments show that \textbf{\textcolor{blue}{CGVQ}} decreases the \ac{bpp} by 20\%$\downarrow$ with respect to our baseline \textbf{\textcolor{red}{\ac{GI}}} \cite{gsImage}, while maintaining on-par visual quality.
\end{abstract}

\begin{CCSXML}
<ccs2012>
 <concept>
  <concept_id>10010147.10010371.10010387</concept_id>
  <concept_desc>Computing methodologies~Image compression</concept_desc>
  <concept_significance>500</concept_significance>
 </concept>
 <concept>
  <concept_id>10010147.10010371.10010352</concept_id>
  <concept_desc>Computing methodologies~Rendering</concept_desc>
  <concept_significance>300</concept_significance>
 </concept>
</ccs2012>
\end{CCSXML}

\ccsdesc[500]{Computing methodologies~Image compression}
\ccsdesc[300]{Computing methodologies~Rendering}

\keywords{Image Compression, 2D Gaussian Splatting, K-Means, Vector Quantizer}

\begin{teaserfigure}
    \centering
    \includegraphics[width=\textwidth]{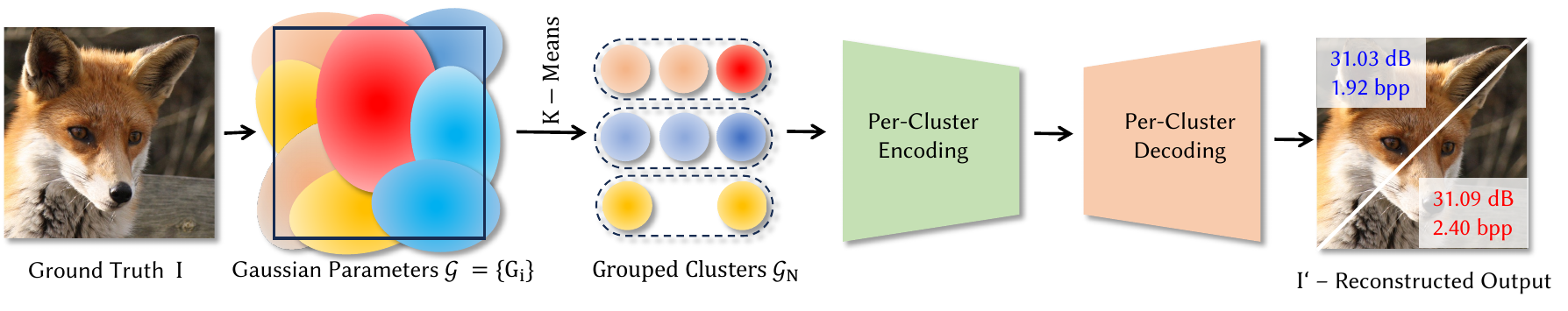}
    \caption{\textbf{Overview of our \textcolor{blue}{\ac{CGVQ}} pipeline for 2D Gaussian-based Image Compression.}
    Given a ground truth image $I$, a set of 2D Gaussian primitives $\mathcal{G} = \{G_i\}$ is first fitted to the image. K-Means clustering partitions $\mathcal{G}$ into $N$ groups $\mathcal{G}_N$ based on appearance and anisotropy similarity. 
    Per-cluster codebooks are then trained and used to 
    encode each group independently. 
    At decoding, each cluster is reconstructed via multiple codebooks lookup and all clusters are composed for rendering, producing the 
    reconstructed output $I'$ with \ac{bpp} at 1.92 and the baseline \textcolor{red}{\textbf{\ac{GI}}} \cite{gsImage} output at \ac{bpp}=2.40 under similar reconstruction fidelity (PSNR $ = 31.0X dB$). Source image \textcopyright{} Animal Faces.}
    \label{fig:teaser}
\end{teaserfigure}


\maketitle

\section{Introduction}
Image compression is essential for real-time rendering \& VR/AR \cite{BeyondBlur} and high-density media storage \cite{5370743}, demanding realistic visual fidelity alongside computational efficiency. 
While traditional codecs like JPEG and PNG provide robust baselines, deep learning architectures \cite{BeyondBlur} have been proposed to further improve the boundaries of rate-distortion performance. 
More Recently, have been posed as a promising alternative for compact image representation \cite{image-gs, gsImage}. 
By modeling local color and structural details with anisotropic Gaussian primitives, Gaussian-based image representation enables flexible, resolution-independent rendering and ultra-fast, GPU-friendly inference \cite{gsImage, 2dgs, sharp}. 
However, this flexibility incurs a storage cost. 
Each 2D Gaussian primitive consists of multiple unconstrained floating-point parameters (e.g., positions, covariance, and color). 
Accurately capturing high-frequency textures requires dense splat allocations, bloating the parameter count and degrading compression efficiency compared to SOTA VAE-based codecs \cite{ivvae, vqvae}. 
To address this, \textcolor{red}{\textbf{\ac{GI}}}~\cite{gsImage} proposed to use Vector Quantization (VQ) and Residual Vector Quantization (RQ) to compress these floating-point attributes into discrete codebooks.
Yet, forcing a capacity-constrained global codebook to map the immense variance of natural image parameters inevitably leads to  quantization errors and visual artifacts. 
\textbf{This limitation demands the need for a representation that preserves an accurate reconstruction of Gaussians while constraining parameter entropy to improve codebook efficiency.}

We propose to leverage a Cluster-based Codebook Quantizing method, \textbf{\textcolor{blue}{\ac{CGVQ}}}.
Based on the principle: for a fixed codebook capacity, narrower parameter distributions are likely to achieve better quantization \cite{huffman}.
We therefore apply K-Means to divide 2D Gaussian primitives into homogeneous clusters and train cluster-specific codebooks for position, rotation-scale, and color, followed by efficient lookup table-based decoding. 
The localized multi-codebook design improves reconstruction quality with only a marginal compression overhead, achieving lower distortion than baseline models within specific bitrate regimes.
Our main contributions are summarized as follows:
\begin{itemize}
    \item We propose \textbf{\textcolor{blue}{\ac{CGVQ}}}, which partitions 2D Gaussian primitives into low-variance groups before quantization.
    \item We show that localized codebooks provide a better compression--fidelity trade-off than a global codebook baseline, improving PSNR/SSIM at comparable bitrate. This highlights the potential of utilizing K-Means clustering as a promising direction for future codebook quantization optimization.
\end{itemize}

Our method \textbf{\textcolor{blue}{{\ac{CGVQ}}}} is inspired and extended based on \textbf{\textcolor{red}{{\ac{GI}}}} \cite{gsImage}. The code is available at \href{https://github.com/complight/Cluster_Guided_Vector_Quantization}{\underline{this repository}}.

\section{Method}
Following \textbf{\textcolor{red}{\ac{GI}}}, we represent the ground truth $I$ as a set of $N$ 2D Gaussian primitives $G_i$:
$
\mathcal{G} = \{G_i\}_{i=1}^{N},\quad G_i = \{\boldsymbol{\mu}_i,\, \mathbf{RS}_i, \mathbf{c}_i\},
$
where $\boldsymbol{\mu}_i \in \mathbb{R}^2$ is the 2D position, $\mathbf{RS}_i \in \mathbb{R}^{2 \times 2}$ is the combined rotation-scale matrix, and $\mathbf{c}_i \in \mathbb{R}^3$ is the weighted color feature \cite{gsImage}. 
In practice, the rotation-scale matrix $RS_i$ is parameterized by two learnable attributes: a scale vector $\mathbf{s}_i \in \mathbb{R}^2$ and a rotation parameter $\mathbf{r}_i \in \mathbb{R}^1$. The matrix used for rendering is then obtained by a deterministic mapping
$
    \mathbf{RS}_i = F(\mathbf{r}_i, \mathbf{s}_i),
$
where $F(\cdot)$ constructs the corresponding rotation-scale matrix from the rotation and scale parameters.

\begin{figure}[h]
    \centering
    \includegraphics[width=\linewidth]{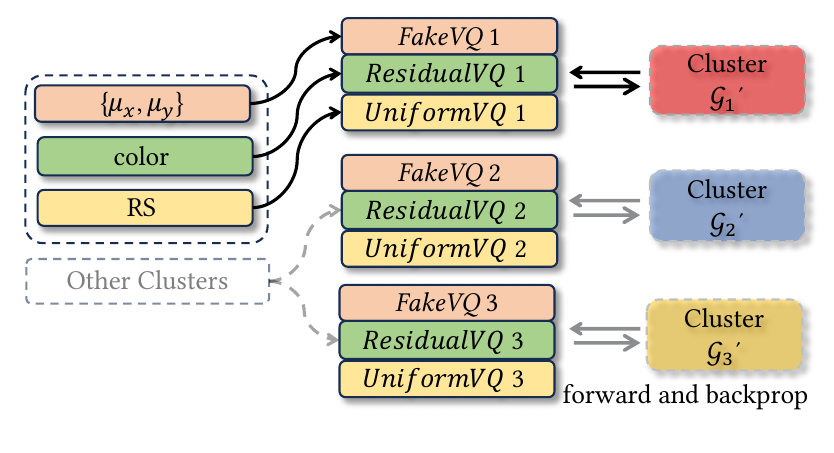}
    \caption{Codebook training and per-cluster compression.}
    \label{fig:codebook_training}
\end{figure}

To enable efficient per-cluster quantization, we group Gaussian primitives by 
feature similarity prior to codebook training.
For each Gaussian primitive $G_i$, we construct a clustering feature vector by concatenating its rotation, scale, and RGB color attributes:
$
    \mathbf{f}_i = \bigl[\,\mathbf{r}_i,\; \mathbf{s}_i,\; \mathbf{c}_i\,\bigr] \in \mathbb{R}^6,
$
where $\mathbf{r}_i \in \mathbb{R}^1$ denotes the rotation parameter, $\mathbf{s}_i \in \mathbb{R}^2$ denotes the two scale parameters, and $\mathbf{c}_i \in \mathbb{R}^3$ denotes the RGB color feature.
The 2D position $\boldsymbol{\mu}_i$ is deliberately excluded, as compression quality shows a high sensitivity to the spatial accuracy.
We then apply K-Means on the normalized features $\{\hat{\mathbf{f}}_i\}$ to 
partition $\mathcal{G}$ into $K$ disjoint clusters by minimizing the total 
intra-cluster variance:
\begin{equation}
    \{\hat{\mathcal{C}_k}\}^K_{k=1}
    \leftarrow
    \argmin_{\{\mathcal{C}_k\}} \sum_{k=1}^{K} \sum_{i \in \mathcal{C}_k} 
    \bigl\|\hat{\mathbf{f}}_i - \boldsymbol{m}_k\bigr\|^2,
\end{equation}
where $\boldsymbol{m}_k$ is the centroid of cluster $\mathcal{C}_k$.
The number of clusters $K$ is selected manually with K between the range of 4 to 16, based on the number of primitives, to keep the balance between reconstruction fidelity and compression efficiency.
Clustering is performed once after the initial training phase converges, and the 
resulting cluster assignment $\{k_i\}$ is shared across all three parameter-specific quantizers, ensuring consistent grouping throughout the entire compression pipeline.
As shown in \refFig{codebook_training}, for each cluster $\mathcal{C}_k$, we train three independent codebooks, one per parameter type, to exploit the distinct statistical properties of each attribute:
For each cluster, we model three components: 
(1) \textbf{position} $\boldsymbol{\mu}$ using 16-bit floating-point QAT with a straight-through FakeVQ estimator, 
(2) \textbf{rotation-scale} $\mathbf{RS}$ using a Uniform Vector Quantization (UQ) codebook, and 
(3) \textbf{color} $\mathbf{c}$ using a Residual Vector Quantization (RQ) codebook for fine-grained residual refinement.
All codebooks are trained end-to-end by minimizing a joint loss combining image reconstruction and vector quantization terms:
$
    \mathcal{L} = \mathcal{L}_{\text{recon}}(\hat{I},\, I) + \mathcal{L}_{\text{vq}}
$
where $\mathcal{L}_{\text{recon}}$ is a weighted combination of a pixel-wise L1 loss in image space and a perceptual L1 loss in feature space, with $\lambda = 0.7$, and $\mathcal{L}_{\text{vq}}$ is the commitment loss that regularizes encoder outputs toward their assigned codebook entries across all active codebooks.

\begin{figure}[h]
    \centering
    \includegraphics[width=\linewidth]{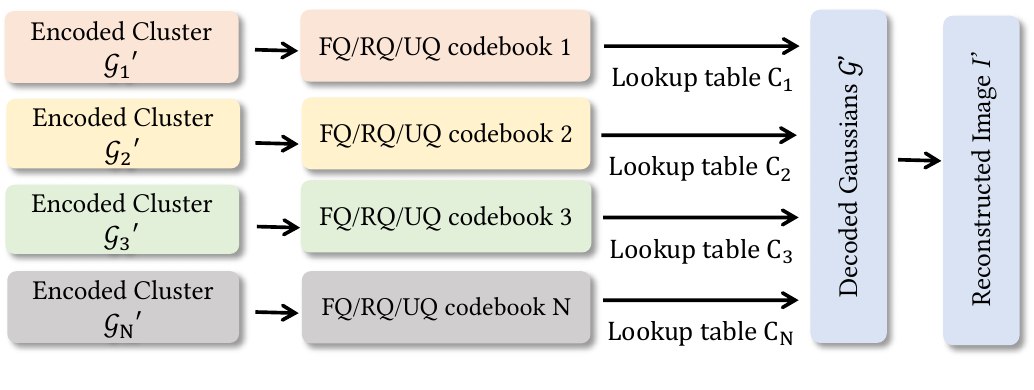}
    \caption{Encoded cluster dequantization pipeline.}
    \label{fig:dequant}
\end{figure}
As illustrated in Figure~\ref{fig:dequant}, each compressed cluster $\mathcal{C}_k$ is decoded independently by looking up its three dedicated codebooks---FP16 for position $\boldsymbol{\mu}$, 
UQ for rotation-scale $\mathbf{RS}$, 
and RQ for color $\mathbf{c}$---using the stored quantization indices $z_i^{\mathbf{RS}}$ and $z_i^{\mathbf{c}}$ to reconstruct each primitive $\hat{G}_i$. 
The decoded primitives from all $K$ clusters are then composed into a single Gaussian set $\mathcal{G}' = \bigcup_{k=1}^{K} \hat{\mathcal{C}}_k$ for rendering.
To further reduce bitrate, we also apply Partial Bits Back Encoding~\cite{gsImage} to the quantization index tables. 

\section{Evaluation}

\begin{figure*}[t]
    \centering
    \includegraphics[width=\linewidth]{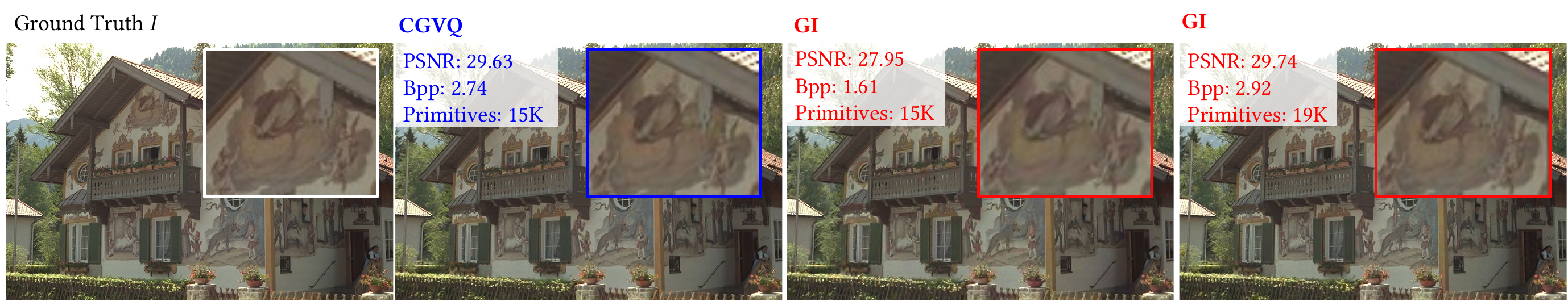}
    \caption{\textbf{Qualitative Comparison of Reconstruction Fidelity.} From left to right: (1) Ground Truth $I$; (2) \textbf{\textcolor{blue}{\ac{CGVQ}}} reconstruction with $15$K primitives; (3) \textbf{\textcolor{red}{\ac{GI}}} baseline with $15$K primitives; (4) \textbf{\textcolor{red}{\ac{GI}}} baseline with $19$K primitives. Source image \textcopyright{} Kodak.}
    \label{fig:visual_compare}
\end{figure*}

\begin{figure}[t]
    \centering
    \includegraphics[width=\linewidth]{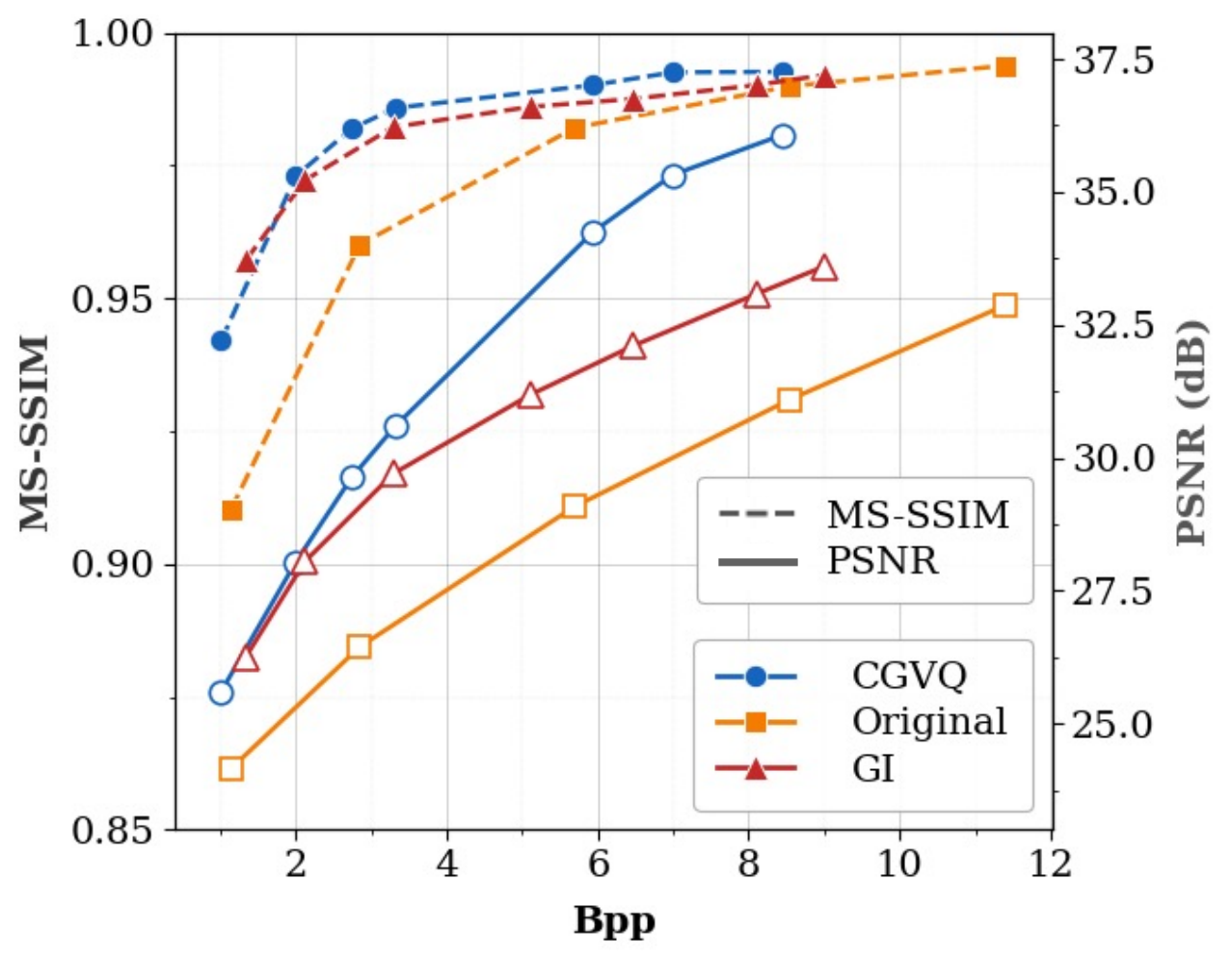}
    \caption{\textbf{Visual Quality:} Rate--distortion performance in terms of PSNR and SSIM versus Bpp on the Kodak dataset, where original refers to uncompressed Gaussian representation.}
    \label{fig:rd_curve}
\end{figure}

% \begin{warfigure}[h]
%     \centering
%     \includegraphics[width=0.75\linewidth]{vis/rd_combined.pdf}
%     \caption{\textbf{Visual Quality: }Rate-Distortion Performance on Kodak Dataset.}
%     \label{fig:rd_curve}
% \end{figure}

In our experiments, $K$ is fixed to 16, which provides a suitable trade-off for scenes containing 10K to 25K Gaussian primitives.
\textbf{Rate-Distortion Efficiency:}
Compared with the \textbf{\textcolor{red}{\ac{GI}}} baseline \cite{gsImage}, \textbf{\textcolor{blue}{\ac{CGVQ}}} consistently demonstrates improved rate-distortion performance on the Kodak dataset. As shown in Fig.~\ref{fig:rd_curve}, \textbf{\textcolor{blue}{\ac{CGVQ}}} achieves higher PSNR and MS-SSIM than \textbf{\textcolor{red}{\ac{GI}}} over the evaluated bitrate range, indicating better reconstruction quality at comparable bitrates. This advantage becomes more evident in the high-quality regime. In particular, for reconstruction quality above approximately $30$ dB, \textbf{\textcolor{blue}{\ac{CGVQ}}} maintains a clear gain in PSNR while also preserving stronger structural similarity, suggesting a more favorable rate-distortion trade-off overall.
\textbf{Parameter Efficiency:}
We further assess parameter efficiency under a fixed Gaussian primitives. As shown in Fig.~\ref{fig:visual_compare}, with \textbf the same {15K} Gaussian primitives, \textbf{\textcolor{blue}{\ac{CGVQ}}} improves PSNR by 1.68 dB$\uparrow$ relative to \textbf{\textcolor{red}{\ac{GI}}}. In addition, \textbf{\textcolor{blue}{\ac{CGVQ}}} at 15K points attains visual quality comparable to \textbf{\textcolor{red}{\ac{GI}}} at 19K points, corresponding to a 21\%$\downarrow$ reduction in primitive count.
The result of \textbf{\textcolor{blue}{\ac{CGVQ}}} stems from the principle that 
\noindent narrower parameter distributions are inherently easier to quantize accurately.
By partitioning 2D Gaussian primitives into $K$ homogeneous clusters based on appearance and anisotropy similarity, we effectively constrain the variance of parameters within each group. This allows our localized multi-codebook design to capture high-frequency details more precisely than global codebook approaches.
\textbf{Performance Trade Off:}
Shown in Table \ref{tab:perf} the number of clusters reduces both encoding and decoding speed, with encoding FPS dropping from $2.91\times10^{-2}$  to $1.9\times10^{-3}$  and decoding FPS from 133.3 to 33.3 as the cluster count increases from 1 to 16. This shows that finer clustering introduces a clear runtime overhead, highlighting a trade-off between compression performance and computational efficiency.
To conclude, \textbf{\textcolor{blue}{\ac{CGVQ}}} improves 2D Gaussian-based image compression by enabling more efficient quantization through cluster-specific codebooks, while introducing a clear trade-off between compression quality and computational efficiency.

\begin{table}[t]
\centering
\caption{\textbf{Performance}: Encoding time and decoding FPS under different cluster settings (primitives 20K)}
\begin{tabular}{lccc}
\toprule
$K$ & PSNR$\uparrow$ & Enc.\ FPS$\uparrow$ & Dec.\ FPS$\uparrow$ \\
\midrule
1  & 29.33                   & \cellcolor{best}$2.91\times10^{-2}$  & \cellcolor{best}133.3  \\
4  & 30.71                   & \cellcolor{second}$5.92\times10^{-3}$  & \cellcolor{second}74.6 \\
8  & \cellcolor{second}30.90 & $3.58\times 10^{-3}$                    & 53.5                   \\
16 & \cellcolor{best}31.18   & $1.9\times 10^{-3}$                    & 33.3                   \\
\bottomrule
\end{tabular}
\label{tab:perf}
\end{table}

\bibliographystyle{ACM-Reference-Format}
\bibliography{sample-base}

\appendix

\end{document}